\documentclass[letterpaper]{article} 
\usepackage{aaai24}  
\usepackage{times}  
\usepackage{helvet}  
\usepackage{courier}  
\usepackage[hyphens]{url}  
\usepackage{graphicx} 
\usepackage{natbib}  
\usepackage{caption} 
\usepackage[ruled,linesnumbered]{algorithm2e}
\usepackage{booktabs}
\usepackage{multicol}
\usepackage{multirow}
\usepackage{amsmath}

\title{Adv-Diffusion: Imperceptible Adversarial Face Identity Attack via Latent Diffusion Model}

\author {
    Decheng Liu\equalcontrib\textsuperscript{\rm 1},
    Xijun Wang\equalcontrib\textsuperscript{\rm 1},
    Chunlei Peng\textsuperscript{\rm 1}\thanks{Corresponding author: Chunlei Peng (clpeng@xidian.edu.cn).},
    Nannan Wang\textsuperscript{\rm 1},
    Ruimin Hu\textsuperscript{\rm 1},
    Xinbo Gao\textsuperscript{\rm 2}
}
\affiliations {
    \textsuperscript{\rm 1}Xidian University, Xi'an, China\\
    \textsuperscript{\rm 2}Chongqing University of Posts and Telecommunications, Chongqing, China\\
}

\begin{document}

\maketitle

\begin{abstract}

Adversarial attacks involve adding perturbations to the source image to cause misclassification by the target model, which demonstrates the potential of attacking face recognition models.
Existing adversarial face image generation methods still can't achieve satisfactory performance because of low transferability and high detectability. 
In this paper, we propose a unified framework Adv-Diffusion that can generate imperceptible adversarial identity perturbations in the latent space but not the raw pixel space, which utilizes strong inpainting capabilities of the latent diffusion model to generate realistic adversarial images.
Specifically, we propose the identity-sensitive conditioned diffusion generative model to generate semantic perturbations in the surroundings.
The designed adaptive strength-based adversarial perturbation algorithm can ensure both attack transferability and stealthiness.
Extensive qualitative and quantitative experiments on the public FFHQ and CelebA-HQ datasets prove the proposed method achieves superior performance compared with the state-of-the-art methods without an extra generative model training process.
The source code is available at https://github.com/kopper-xdu/Adv-Diffusion.
\end{abstract}
\begin{figure*}
    \centering
    \includegraphics[width=0.7\textwidth]{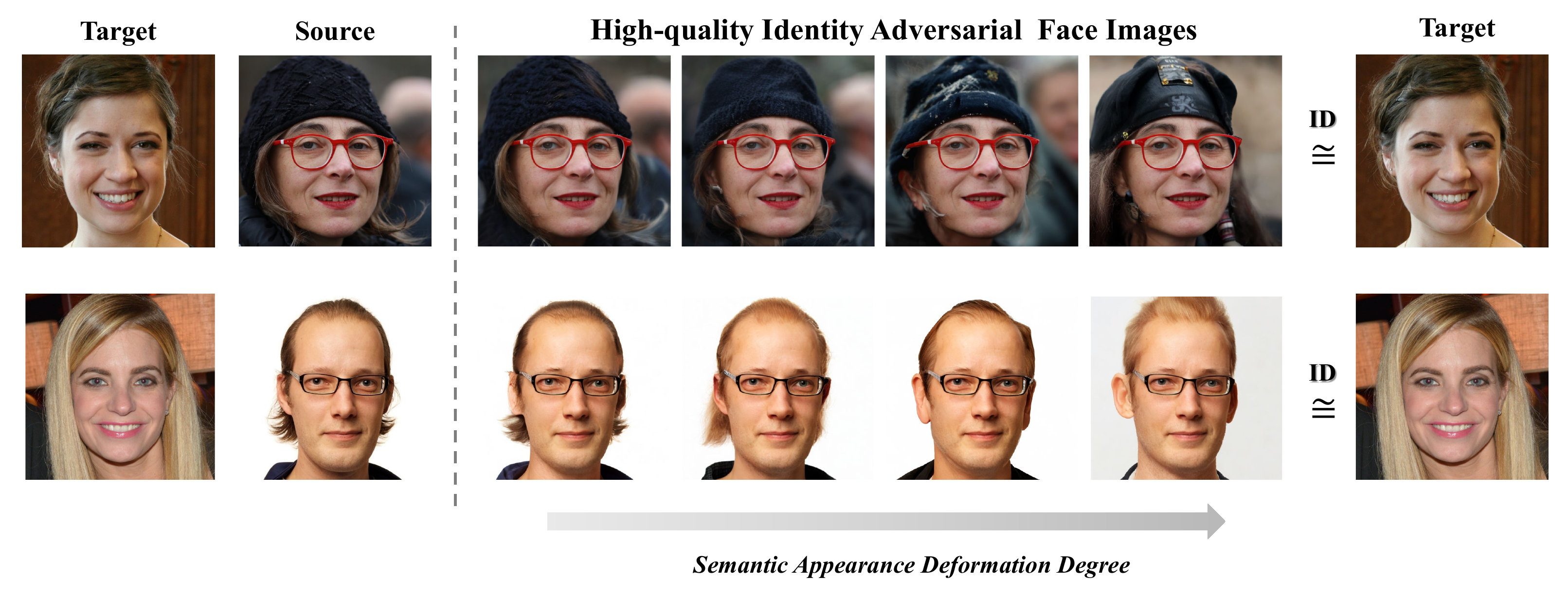}
    \caption{Illustrations of our proposed imperceptible adversarial face identity attack task, which can generate high-quality faces with target identities and similar source appearance information.}
    \label{figure1}
\end{figure*}
\section{Introduction}

With the development of deep learning technology, face recognition systems have been applied in more and more real-world scenarios, which also brings increasing security risks of biometrics in the meantime.
Recent works find that these deep neural network-based models are vulnerable to adversarial perturbations added in the original clean images \cite{fgsm,pgd}.
These well-designed perturbations added images, also called \emph{adversarial examples}, demonstrate the strong potential of attacking against existing state-of-the-art face recognition models \cite{adv-faces,adv-glasses}, even evaluating black-box attacking scenarios \cite{adv-attribute,adv-hat}. 
Thus, exploring craft adversarial examples in face recognition is vital and significant in the field of biometrics and economic security.

Existing adversarial attacks on face recognition methods are roughly grouped into three categories:
gradient-based methods, patch-based methods and stealthy-based methods.
The goal of adversarial face examples is to successfully attack target models with strong transferability, and generate high-quality adversarial images with inconspicuous perturbations.
Gradient-based adversarial examples methods are early explorations to add $L_p$ bounded perturbations directly to source images, which could bring in arbitrary predictions when testing \cite{pgd,mifgsm}. 
By controlling the perturbation boundary, pixel-level attacking clues seem imperceptible and make it more feasible to invade deployed face recognition systems.
However, some related studies have found that gradient-based methods are vulnerable to variant lighting conditions \cite{xiao2018spatially} and perform poorly in black-box attacking evaluations \cite{gen-ap}.  
Patch-based methods \cite{adv-glasses,adv-hat} aim to generate adversarial local face patches to protect identity information in physical-world scenarios.
These existing synthesized craft adversarial patches usually have specific color and texture patterns, which are more easily distinguished and fail to be stealthy. 
Formerly mentioned kinds of adversarial face examples methods are generating suitable perturbations in raw pixel space, ignoring the specific properties of face images inherently.
\cite{qiu2019semanticadv} first explored learning adversarial face examples with semantic appearance, which could help to make these images visually imperceptible. 
\cite{amt-gan} leveraged the makeup transfer generative network to generate better visual-quality adversarial faces.
Recently, \cite{adv-attribute} has focused on synthesizing adversarial faces with edited attributes from reference faces to improve the stealthiness of attacking information.
However, noting that not any reference face images would provide suitable and precise semantic information (e.g. attribute text \cite{qiu2019semanticadv}, semantic clues \cite{adv-attribute}) as guidance, and 
\emph{it is essential to learn adversarial semantic appearance automatically with high attack capabilities.}

To address these issues, we propose a novel imperceptible adversarial face identity attack via latent diffusion model (Adv-Diffusion), which can generate adversarial perturbation in the latent space but not the raw pixel space and result in less perceptive perturbations surrounding the identity-sensitive regions as shown in Figure \ref{figure1}.
Specifically, we leverage the latent diffusion model to construct the latent space for adversarial semantic perturbations. 
Because the diffusion model could offer excellent inductive bias for spatial data, and the learned latent space is perceptually equivalent to the raw pixel space \cite{ldm}.  
To improve the key stealthiness performance, we design the identity-agnostic conditioned diffusion generative module.
The pre-trained parsing model \cite{luo2012hierarchical,liu2020new} is utilized to disentangle the face identity-sensitive regions (e.g. eyes, nose, mouth, etc.) and the face identity-agnostic region (e.g. hairstyle, decoration, background, etc.) by masking operation from a cognitive psychology perspective.
Moreover, the identity-sensitive regions are regarded as the condition of the latent diffusion model when restoring adversarial face images.
Additionally, the semantical adversarial perturbations are learned to add to the latent embedding through the adversarial attacking against face recognition.
Noting that the proposed method is without any training data or complex deep network architectures, which is benefitted from the specific properties of the latent diffusion model.

The main contributions of our paper are summarized as follows:
\begin{itemize}
\item We propose a novel unified adversarial face image generation pipeline attacking face recognition in the latent space but not the raw pixel space, which could automatically learn effective adversarial semantic appearance with high attack capabilities and low perceptibility.
\item We further propose the identity-sensitive conditioned diffusion generative module to guarantee the most adversarial appearance surrounding the identity-sensitive region, and the designed adaptive strength-based semantical adversarial perturbation is designed to ensure attack transferability and stealthiness.
\item Experimental results on the public FFHQ and CelebA-HQ datasets illustrate the superior performance of the proposed Adv-Diffusion compared with the state-of-the-art adversarial face image generation methods. Meanwhile, the proposed method is without any training data or complex deep network architectures, which makes it easy to be deployed in real-world scenarios.
\end{itemize}

\section{Related Work}

\begin{figure*}
    \centering
    \includegraphics[width=0.8\textwidth]{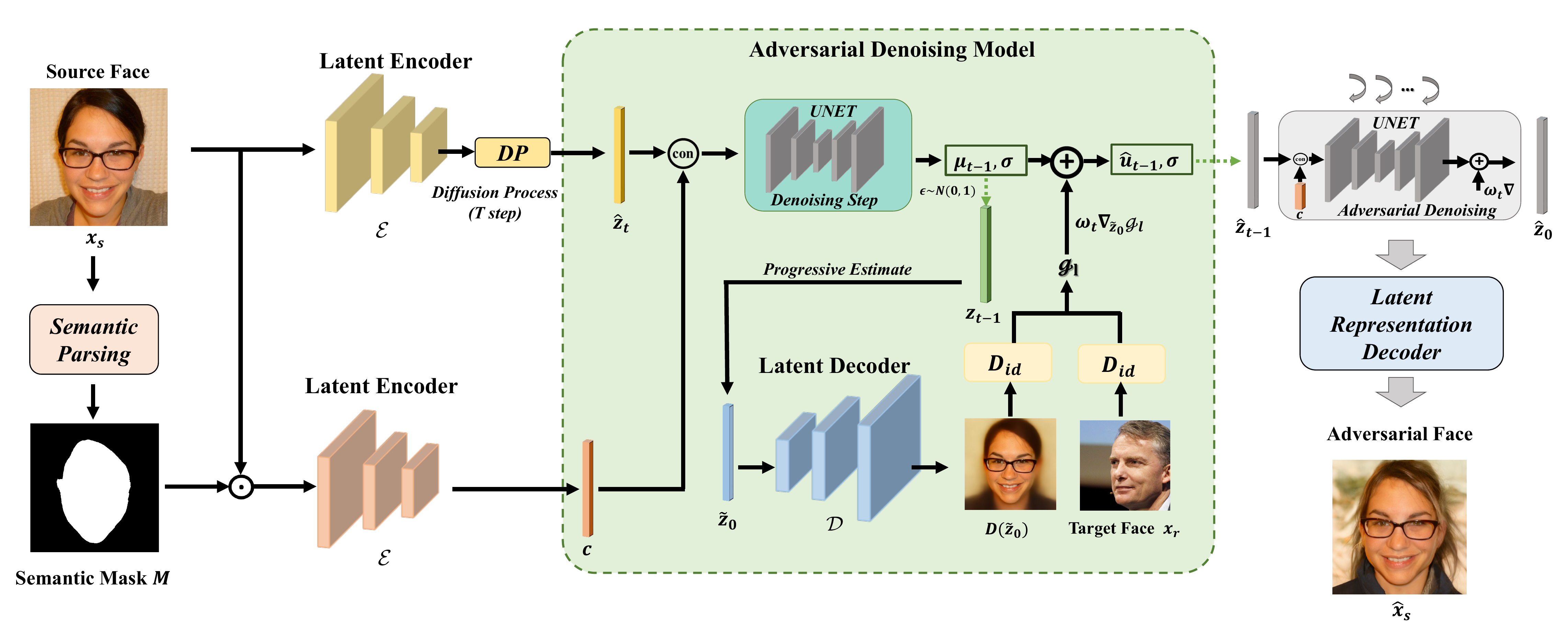}
    \caption{The framework of the proposed imperceptible adversarial face identity attack via latent diffusion model.}
    \label{figure2}
\end{figure*}

\subsection{Adversarial Attack on Face Recognition Model}

Adversarial attacks on face recognition model is a growing field of research, and many studies have been conducted to evaluate the robustness of advanced face recognition models against adversarial attacks. In this paragraph, we review the representative attack methods on the face recognition model in
three categories: gradient-based, patch-based and stealthy-based methods.
\paragraph{Gradient-based method.}

Gradient-based methods are one of the most widely used attack techniques in deep learning models.
And several gradient-based methods have been proposed for adversarial attacks, including Fast Gradient Sign Method (FGSM)~\citep{fgsm}, Projected Gradient Descent (PGD) \citep{pgd} and Momentum-based Iterative FGSM (MI-FGSM) \citep{mifgsm}. 
FGSM is a fast and effective method for generating adversarial examples, where the perturbation is added to the input in the direction of the sign of the gradient. PGD 
is an iterative variant of FGSM that applies multiple small perturbations to the input and uses a projected gradient descent algorithm to ensure that the perturbation remains within a specified range. However, gradient-based methods are mostly applied in the white-box setting and are perceptible to humans.

\paragraph{Stealthy-based method.}

To overcome these limitations, recent studies have focused on developing imperceptible attacks with high transferability. For example, \citep{adv-makeup} proposed Adv-Makeup
aims to achieve an implementable physical attack with high transferability by focusing on the makeup worn by the target individual. This approach is not only effective in digital conditions but also in physical scenarios, as the synthesized makeup is imperceptible to the human eye. Similarly, \citep{amt-gan} proposed AMT-GAN, which also focuses on makeup transfer attack and has more transfer area. These above methods focus on pixel space control, recent work \citep{adv-attribute} proposed Adv-Attribute to leverage Stylegan model \citep{stylegan} to attack the latent vector of the input image when preserving the identity of the input image. 
\citep{barattin2023attribute} directly optimizes the latent representation of images in the latent space of a pre-trained GAN, which preserves the original facial attributes and attacks the recognition model successfully.

\paragraph{Patch-based method.}

The patch-based method is mainly focused on the physical attack in real-world scenarios.
For instance, Adv-Glasses \citep{adv-glasses} proposed a patch-based attack using optimization-based methods to add perturbations to the eyeglass region, while Adv-Hat \citep{adv-hat} generated adversarial attacks over wearing hats. To improve attack ability when the attacker has limited accessibility to the target models, \citep{gen-ap} extend the existing transfer-based attack techniques to generate transferable adversarial patches by singing face-like features as adversarial perturbations through optimization on the manifold.

\subsection{Diffusion Model Variants}

Diffusion models, also known as diffusion probabilistic models (DPM), have recently emerged as a powerful generative modeling technique for image and video data. 
DPMs have shown state-of-the-art performance in sample quality. The core of DPMs is a diffusion process that evolves a simple initial distribution to a more complex target distribution by applying a sequence of diffusion steps. 
And there are several variants of DPMs.
\cite{score-based} proposed a score-based generative model which learns the score function of data distribution. 
And continuous-time DPM defined stochastic differential equations to describe the diffusion and sample process of DPM.
DiffPure \cite{DiffPure} leveraged the denoising ability of DPM for defending against adversarial attacks recently. 
\cite{ldm} proposed latent diffusion model, which firstly learns an autoencoder to map image data to latent space, and further leverages the diffusion model to learn the latent space distribution. The latent diffusion model has more flexible conditional generation ability than pure DPM, such as inpainting, text-based image generation and image translation. 
Extensive analysis has proved the latent diffusion model can generate more realistic and high-fidelity results.
Inspired by these successes, we also leverage the latent diffusion model to learn the latent space, but with a different target which is to generate identity adversarial face images with imperceptible perturbations.

\section{Methodology}

\paragraph{Problem Definition.} 

The goal of the adversarial face identity attack is to mislead existing advanced face recognition with adversarial perturbations.
However, most former works mainly focus on learning $L_p$ bounded perturbations in the raw pixel space to make less detectability. 
Considering the specific properties of adversarial face images, we give a broader definition as follows.
The general objective of targeted adversarial face images generation is defined as:
\begin{equation}
\label{eq1}
arg \max_{\hat{x}_s} F\left(x_{r}, \hat{x}_s\right).
\end{equation}
Here $F(\cdot)$ refers to the identity similarity of these paired faces, where $x_s$ and $x_r$ denote 
the source image and target image respectively.
The goal of our method is to learn a strong generative model, denoted as $G(\cdot)$, to generate the adversarial face image $\hat{x}_s$ with the condition of the target face $x_s$.
Noting that the formulation can be easily extended to the untargeted adversarial face image generation application.
\emph{The criteria of adversarial face image should contain both strong transferability on recognition evaluation and less perceptive perturbations. 
}

\subsection{Preliminaries: Latent Diffusion Model}

In this section, we mainly introduce the conditional latent diffusion model (LDM).
The diffusion model can generate high-quality images
by reversing a nosing process iteratively.
The latent diffusion model \cite{ldm} learns a compressed latent space of lower dimensions for decreasing computing cost.
We firstly define an autoencoder to encode input image $x$ into a latent code $z = \mathcal{E}\left(x\right)$ and decode $z$ to $x = \mathcal{D}\left(z\right)$. 
During the forward stage, the T-step forward diffusion process aims to transform latent code $z$ gradually to match the Gaussian distribution: $z_T \sim \mathcal{N}(0,1)$.
The forward process can be defined as 
$
  q\left(z_t \mid z_{t-1}\right) := \mathcal{N}\left(z_t ; \sqrt{1-\beta_t} z_{t-1}, \beta_t \mathbf{I}\right),
$
where $\beta_t \in (0,1]$ is the hyper-parameter.
Following, the reverse process of the denoising step can be expressed as follows:
\begin{equation}
\label{eq2}
  p_\theta\left(z_{t-1} \mid z_t\right) := \mathcal{N}\left(z_{t-1} ; \mu_\theta\left(z_t, t\right), \Sigma_\theta\left(z_t, t\right)\right),
\end{equation}

where $\mu_\theta$ and $\Sigma_\theta$ separately mean the mean and standard deviation, where
$
\mu_\theta\left(z_t, t\right) = \frac{1}{\sqrt{\alpha_t}}\left(z_t-\frac{1-\alpha_t}{\sqrt{1-\bar{\alpha}_t}} \epsilon_\theta\left(z_t, t\right)\right).
$
During the inference stage, we randomly initialize the latent code $z_T \sim \mathcal{N}(0,1)$,
and then gradually update it until $z_0$ with the mentioned reverse process.
Finally, the synthesized high-quality face image $x_0$ could be generated by the latent representation decoder.

\subsection{Identity-sensitive Conditioned Diffusion Generative Model}
\begin{figure}
  \centering
  \includegraphics[width=0.45\textwidth]{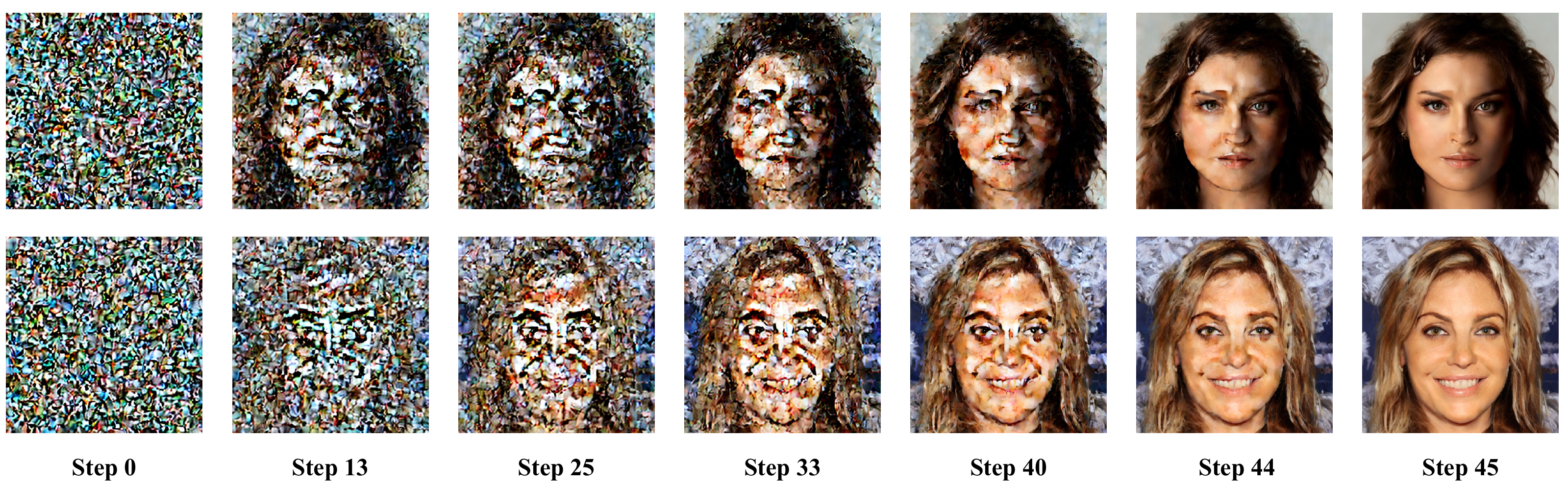}
  \caption{The reverse process with adversarial perturbation in the proposed method.}
\label{figure_reverse}
\end{figure}

Motivated by the perspective of cognitive psychology, most identity-discriminative information concentrates on identity-sensitive regions, e.g. eyes, nose, cheek, etc.
The identity-agnostic regions (like hairstyle, decoration, background, etc) contain less discriminative information for human recognition but can be captured in the face recognition model (shown in Figure \ref{figure2}).
Thus, we leverage the pre-trained face parsing model to calculate the face region binary mask $M$.
Then, we calculate the identity-sensitive region as $x_m = x_s \odot \left(1-M\right)$.
Following, we leverage the strong inpainting capabilities of LDM to guarantee the most adversarial semantic appearance perturbations surrounding the identity-sensitive region.
We leverage the pre-trained latent encoder to map the identity-sensitive region into the latent space $c = \mathcal{E}\left(x_m\right)$.
We also calculate the initial value of $\hat{z}_T$ in the mentioned forward diffusion process with $T$ steps by using $x_s$ as the input image.
The $\hat{z}_T$ is concatenated with $c$ to obtain the input of the UNet model.
Then, with the similar inpainting inference procedure in \cite{ldm}, the reverse process with the designed adversarial denoising model can be expressed as:
\begin{align}
\label{eq3}
p_\theta\left(z_{t-1} \mid \hat{z}_t, c\right) &:= \mathcal{N}\left(z_{t-1}; \mu_\theta\left(\hat{z}_t, t, c\right), \Sigma_\theta\left(\hat{z}_t, t, c\right)\right) \\
\hat{z}_{t-1} &:= z_{t-1} + \mathcal{G}_t
\end{align}
where $\mathcal{G}_t$ is the designed adversarial perturbations described in the following. Figure \ref{figure_reverse} demonstrates the change of $\mathcal{D}(\hat{z}_{t-1})$ during the reverse process.
Finally, the restored adversarial face image $\hat{x}_s$ can be calculated by the latent representation decoder as $\hat{x}_s = \mathcal{D}(\hat{z}_0)$.

\subsection{Adaptive Strength based Semantical Adversarial Perturbation}

To generate less perceptible perturbations in the identity-sensitive regions, we add the learned perturbations in the latent space but not the raw pixel space.
We assume that the mentioned identity-sensitive conditioned diffusion model mainly focuses on generating suitable visual information surrounding the central regions and tries to maintain the identity-sensitive regions because of the condition of $c$ and the inpainting ability of LDM.
To leverage the above ability, we add semantic adversarial perturbations to the latent space in the reverse inpainting process.

Because the latent space is perceptually equivalent to the raw pixel space, we design the added perturbations in the latent space that would bring in diverse semantic appearances.
Specifically, we design the simple gradient-based adversarial sample algorithm to generate adversarial semantic perturbations.
Here we utilized the face recognition model as the target model and the adaptive strength-based adversarial perturbation is calculated as follows:
\begin{equation}
\label{eq4}
\mathcal{G}_t =  w_t\nabla_{\widetilde{z}_0} F\left(D\left(\widetilde{z}_0\right), x_r\right),
\end{equation}
where $w_t=s\Sigma_\theta\left(\hat{z}_t, t, c\right)$, $s$ is the hyper-parameter to control attack strength.
$\widetilde{z}_0$ is an approximate result predicted from $z_{t-1}$ as $\widetilde{z}_0 = \frac{1}{\sqrt{\bar{\alpha}_t}}\left(z_{t-1}-\sqrt{1-\bar{\alpha}_t} \epsilon_\theta(z_{t-1}, t)\right)$ with similar estimation \cite{DDPM}.
$\alpha$ is also a hyperparameter mentioned before.
The details of the parameters analysis are shown in the following section.
It noted that when the reverse step increases, the attack strength $w_t$ will decrease adaptively because of $\Sigma_\theta\left(\hat{z}_t, t, c\right)$.
The designed adaptive attack strength strategy would help improve image quality.
More algorithm details are shown in Algorithm 1.

\begin{algorithm}[!h]
  \SetAlgoLined
  \KwIn{source image $x_s$, target image $x_r$}
  \KwOut{adversarial image $\hat{x}_s$}
  Initialization pretrained LDM, set s, T value\; 
  $z_0 = \mathcal{E}(x_s), M = f(x_s), x_m = x_s \odot (1 - M)$\;
  $\hat{z}_{T} = \sqrt{\bar{\alpha}_T}z_0 + \sqrt{1 - \bar{\alpha}_T}\epsilon, \epsilon \sim \mathcal{N}(0,1)$\;
  $c = \mathcal{E}\left(x_m\right)$\;
  \For{all t from T to 1}
  {
    $z_{t-1} \leftarrow \mu_\theta\left(\hat{z}_t, t, c\right) + \epsilon\Sigma_\theta\left(\hat{z}_t, t, c\right), \epsilon \sim \mathcal{N}(0,1)$\;
    $\widetilde{z}_0 \leftarrow \frac{1}{\sqrt{\bar{\alpha}_t}}\left({z}_{t-1}-\sqrt{1-\bar{\alpha}_t} \epsilon_\theta(z_{t-1}, t)\right)$ \;
    $\mathcal{G}_t \leftarrow w_t\nabla_{\widetilde{z}_0} F\left(\mathcal{D}\left(\widetilde{z}_0\right), x_r\right)$ \;
    $\hat{z}_{t-1} \leftarrow z_{t-1} + \mathcal{G}_t$ \;
    
  }
  \caption{Adv-Diffusion}
\end{algorithm}

\section{Experiments}

\begin{figure*}[htbp]
  \centering
  \includegraphics[width=0.68\textwidth]{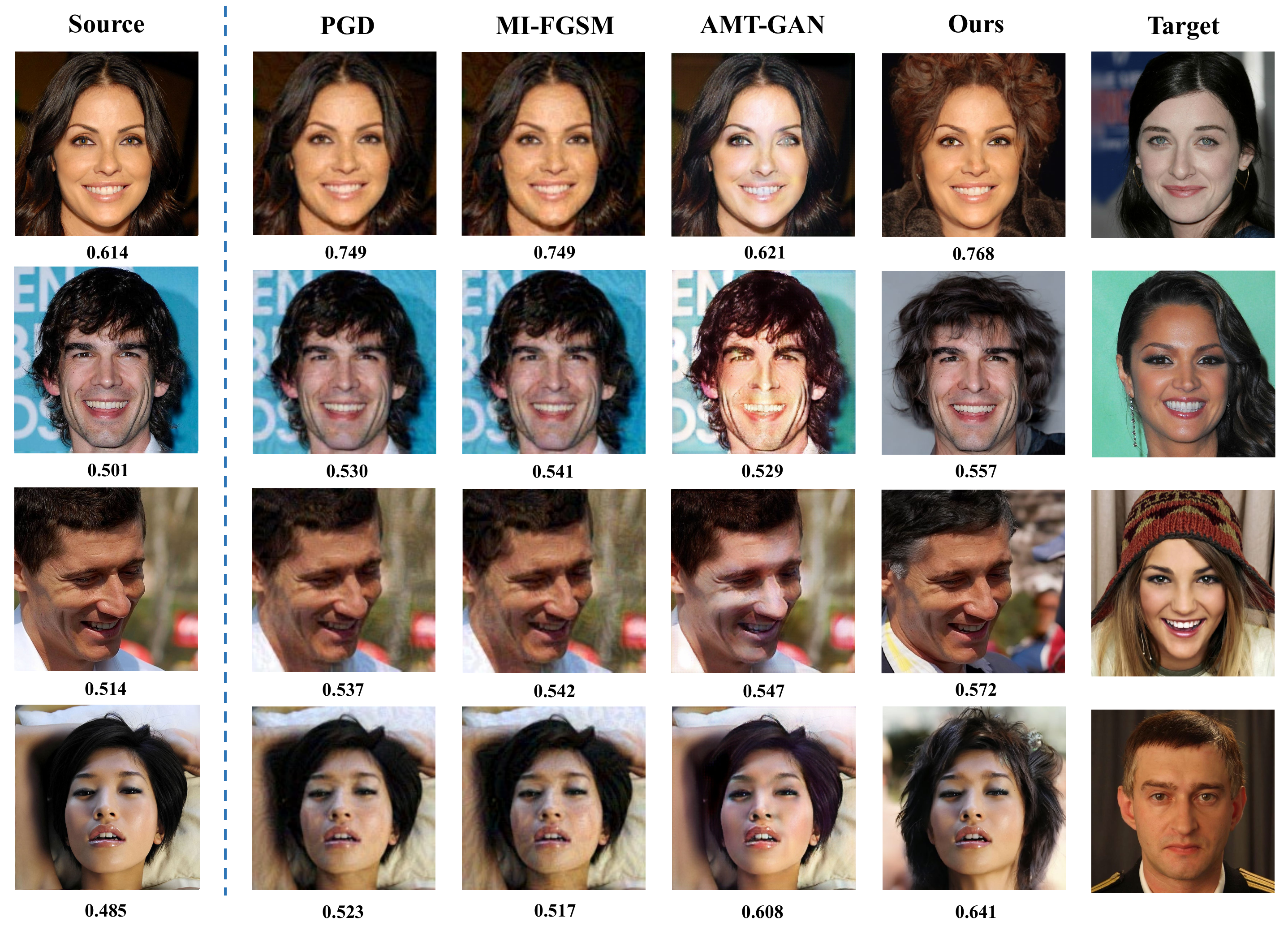}
  \caption{Comparison Results of generated adversarial samples when T is set as 1000. The number below the images means the calculated similarity score with the target face by the refined recognition model. Noting that the identity similarity score varies from 0 to 1.}
    \label{comparison_fig}
\end{figure*}

\subsection{Experimental Setting}

\paragraph{Datasets.}

Following similar protocols \cite{adv-attribute}, we use two publicly available face datasets for evaluation: 
(1) FFHQ is a widely used high-quality face dataset \cite{stylegan}, which contains almost 70,000 
high-quality face images with ${1024\times1024}$ resolution. 
(2) CelebA-HQ is a high-quality face dataset \cite{CelebAHQ} constructed based on the CelebA dataset, which contains almost 30,000 face images with ${512\times512}$ resolution. 
In the evaluation stage, we randomly select 1000 images 
with different identities as the source images for both datasets. 
In particular, we also select 5 additional images as target images for each dataset, and randomly divide the 1000 source images into 5 groups, each corresponding to a different target image.


\paragraph{Implementation details.}
We utilize the pre-trained latent encoder and decoder networks, which are inspired by the open-source stable diffusion work.
For experimental settings, we set 45 steps to generate adversarial samples.
And we set s = 300 by default. 
Additionally, the face semantic parsing model architecture is based on the PyTorch implementation of EHANet \cite{luo2020ehanet}.
The face parsing model can segment holistic faces into 19 semantic regions, including face, hair, background, etc.
We conduct experiments on RTX 3090 GPU. 

\paragraph{Evaluation metrics.}
We evaluate our approach by quantifying the 
attack performance and image quality separately.
The attack success rate (ASR) \cite{adv-attribute} is utilized to evaluate the attack performance of adversarial example algorithms.
The value of $\tau$ is set as 0.01 FAR (False Acceptance Rate) with the same setting in  \cite{amt-gan}. More details are shown in the supplement.
For evaluating the generated image quality, we use the Frechet Inception Distance (FID), Peak Signal-to-Noise Ratio (PSNR) and Structural Similarity (SSIM) as quality metrics in the following experiments.

\subsection{Comparison Results}

We first use the ASR metric to compare the proposed Adv-Diffusion with SOTA methods in black-box attack scenarios.
The comparison results prove the proposed algorithm's high transferability in recognition attack tasks.
Additionally, we evaluate the generated image quality compared with these mentioned SOTA methods to prove the evident imperceptibility of the proposed algorithm.
The following experimental results prove our proposed method can achieve superior performance in both attacking transferability and imperceptibility.

\paragraph{Attacks on black-box model.}

To evaluate the attack performance of the proposed method, we selected four commonly used 
face recognition models: IR152 \cite{he2016deep} , IRSE50 \cite{hu2018squeeze}, FaceNet \cite{schroff2015facenet}, and MobileFace \cite{deng2019arcface} as target models following \cite{amt-gan}.
It is noting that these face recognition models are pre-trained in large-scale face datasets, and all achieve satisfactory recognition performance.
When one of these models is selected as a black-box target model, 
the other three models are used as white-box target models for generating attacking adversarial examples.
We selected three categories of attack methods as our competitors: gradient-based methods, patch-based methods, and stealthy-based methods. 
For gradient-based methods, we set the perturbation strength value to 8/255. 
For other methods, we followed the official parameter settings. More details on the parameter setting can be found in the supplementary materials.
Table \ref{ASR_results} shows the attack performance comparison results on CelebA-HQ and FFHQ datasets. 
Noting that asterisk notation means the result is derived from source paper.
Experimental results prove the proposed method achieves the highest attacking performance compared with the SOTA methods.

\begin{table}
\caption{Generated image quality comparison results with several metrics when T is set as 50.}
\label{img_quality}
  \centering
    \begin{tabular}{*{4}{c}}
      \toprule
      Dataset & \multicolumn{3}{c}{CelebA-HQ} \\
      \cmidrule(lr){1-1} \cmidrule(lr){2-4}
      Metric & FID ($\downarrow $)  & PSNR ($\uparrow $) & SSIM ($\uparrow $) \\
      \midrule
      FGSM & 108.99 & 27.60 & \underline{0.81} \\
      PGD  & 79.92 & 27.96 & \textbf{0.85} \\
      MI-FGSM & 79.42 & \underline{28.85} & 0.82\\
      AMT-GAN & \underline{22.57} & 9.31 & 0.387\\
      Adv-Attribute* & 68.52 & - & -\\
      \midrule
      Adv-Diffusion & \textbf{15.51} & \textbf{29.01} & 0.80 \\
      \bottomrule
    \end{tabular}

\end{table}

\begin{table*}
\caption{The comparison of experimental results with ASR metric when T is set as 50.}
\label{ASR_results}
  \centering
   \renewcommand\arraystretch{1.1}        
   \resizebox{\linewidth}{!}{
    \begin{tabular}{*{10}{c}}
      \toprule
      \multirow{2}*{Method} &  Dataset & \multicolumn{4}{c}{CelebA-HQ} & \multicolumn{4}{c}{FFHQ} \\
      \cmidrule(lr){2-2} \cmidrule(lr){3-6} \cmidrule(lr){7-10}
      & Target Model & IR152 & IRSE50 & FaceNet & MobileFace & IR152 & IRSE50 & FaceNet & MobileFace \\
      \midrule
      Clean & - & 4.72 & 5.40 & 0.80 & 14.84 & 3.56 & 4.94 & 1.53 & 9.60\\
      \midrule
      \multirow{3}*{Gradient-based} & FGSM & 11.52 & 47.13 & 1.22 & 54.75 & 10.00 & 51.01 & 3.95 & 53.26 \\
      & PGD & 41.80 & 67.09 & 20.62 & 59.51 & 38.84 & 73.84 & 19.34 & 62.32 \\
      & MIFGSM & \underline{44.81} & \underline{77.05} & 27.81 & \underline{65.30} & 44.21 & \underline{81.00} & 24.54 & \underline{67.84} \\
      \midrule
      \multirow{2}*{Patch-based} & Adv-Hat* & 2.50 & - & 4.70 & 8.40 & 13.60 & -& 4.80 & 3.10 \\
      & Gen-AP* & 19.50 &  - & 15.80 & 24.40 & 12.00 & -& 8.20 & 19.90 \\
      \midrule
      \multirow{2}*{Stealthy-based}
      & Adv-Attribute* & 44.30 & - & \underline{31.80} &50.20  & \underline{46.30} & - & \underline{31.90} & 49.90\\
      & AMT-GAN & 10.38 & 55.28 & 5.08 & 46.8 & 14.35 & 60.21 & 7.23 & 45.12\\
      \midrule
      Ours & Adv-Diffusion & \textbf{53.33} & \textbf{84.11} & \textbf{36.84} & \textbf{73.02} & \textbf{53.42} & \textbf{82.91} & \textbf{32.21} & \textbf{69.25} \\
      \bottomrule
    \end{tabular}}

\end{table*}

\paragraph{Image quality assessment.}
To prove the imperceptibility of the proposed adversarial attack method, we further evaluate the adversarial image quality compared with other adversarial attack algorithms.
Here we choose common image quality metrics, like Frechet Inception Distance (FID) \cite{heusel2017gans}, Peak Signal-to-Noise Ratio (PSNR) and Structural Similarity (SSIM).
The better-generated image quality always means the lower imperceptibility of adversarial attack perturbations.
Table \ref{img_quality} quantifies the adversarial image quality results of our method compared with the SOTA methods.
Moreover, we have also evaluated the generated adversarial samples from a qualitative perspective. 
Figure \ref{comparison_fig} compares the adversarial samples generated by our method and other methods.
The results prove the proposed methods can maintain better stealthiness of perturbation.
Furthermore, we conducted a series of user studies to evaluate the performance in the supplement.

\begin{figure}[htbp]
  \centering
  \includegraphics[width=0.70\linewidth]{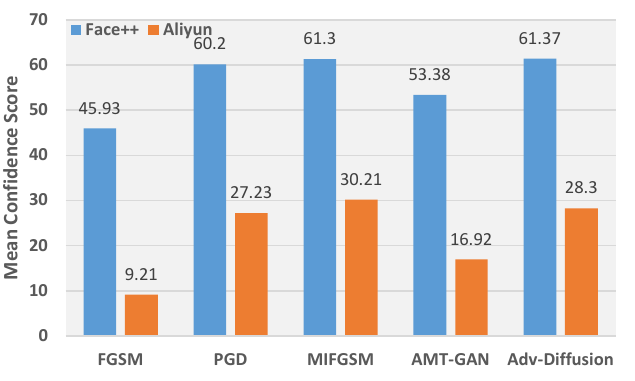}
  \caption{Mean confidence scores calculated from Face++ and Aliyun, which illustrate the possibility of the generated adversarial face image having the same identity as the target image.}
\label{figure5}
\end{figure}

\paragraph{Attacks on commercial APIs.}

To further prove the effectiveness and transferability of our method in real-world scenarios, we conduct comparison experiments on two commercial APIs. 
We choose Face++ and Aliyun API as the target models and comparison experiment results on the CelebA-HQ dataset are shown.
Figure \ref{figure5} shows the proposed method outperforms SOTA methods on Face++ API and achieves competitive performance on Aliyun API.

\subsection{Ablation Study}
In this section, we explore the effect of two key components of the proposed method.
The first component is the adaptive strength utilized to control the magnitude of adversarial perturbations for better image quality.
The second component is the masked image condition in the designed attack model to improve stealthiness.
We separately conduct quantitive and qualitative evaluation experiments to prove the effectiveness of the proposed method.

\begin{table}[h]
  \caption{Ablation study experimental results with several metrics.}
  \centering
    \begin{tabular}{*{7}{c}}
      \toprule
      Method & ASR ($\uparrow $)  & FID ($\downarrow$)  & PSNR ($\uparrow $) \\
      \midrule
      Adv-Diffusion & 53.43 & \textbf{15.34} & \textbf{22.01}\\
      w/o Adaptive Strength & \textbf{71.11} & 62.37 & 12.62 \\
      w/o Mask & 59.11 & 55.23 & 13.22\\
      \bottomrule
    \end{tabular}
    \label{ablation_table}
\end{table}

\paragraph{Quantitative evaluation.}
We quantitatively investigate the impact of the adaptive strength strategy and the mask-conditioned generative model with ASR, FID and PSNR metrics.
The experiments are conducted on the CelebA-HQ dataset using IR152 as the target model, and T is set as 50 for the following experiments. 
Table \ref{ablation_table} proves the adaptive strength strategy can help achieve high attacking performance and maintain good image quality.

\begin{figure}[h!]
  \centering
  \includegraphics[scale=0.26]{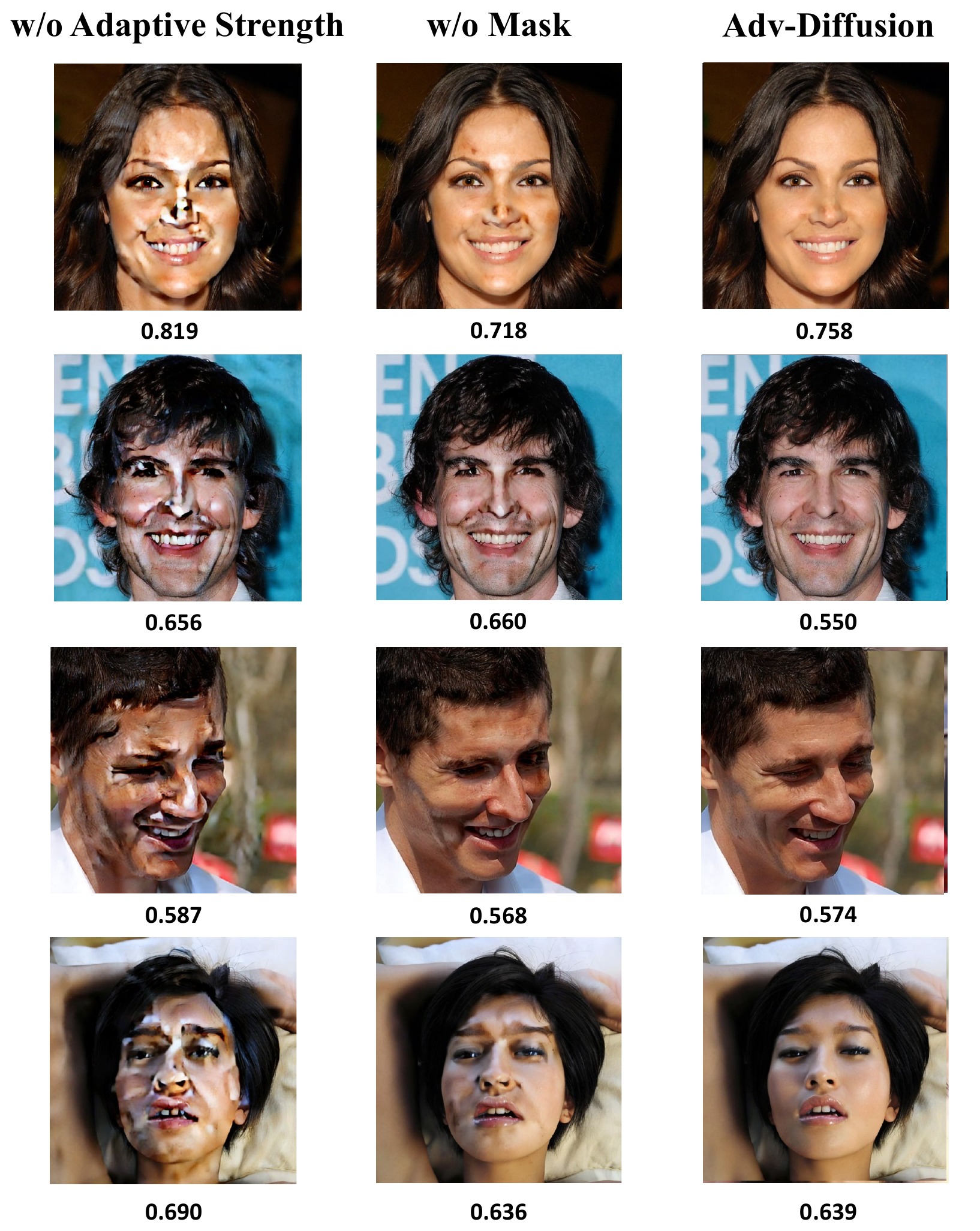}
  \caption{Ablation study results of generated adversarial samples when T is set as 50. The number below the images means the calculated similarity score with the same target face by the refined recognition model.}
\label{figure6}
\end{figure}

\paragraph{Qualitative evaluation.}
For the convenience of analysis, we further conduct the qualitative evaluation as shown in Figure \ref{figure6}. 
It can be observed that the designed adaptive strength and masked image-conditioned module can greatly enhance the quality of generated images while maintaining similar attack performance.

\begin{figure}[h]
  \centering
    \includegraphics[width=0.3\textwidth]{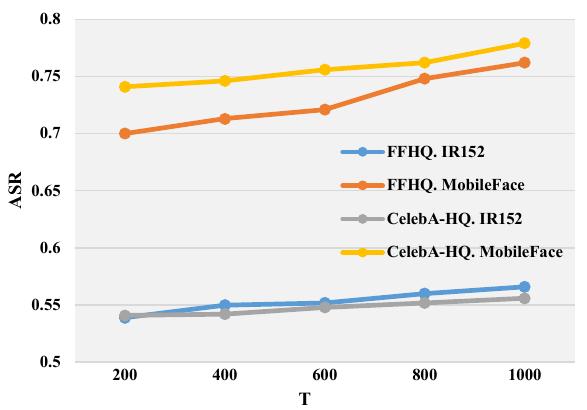}
      \caption{The ASR results with different T values.}
\label{figure7}
\end{figure}

\subsection{Parameter Analysis}

We explore the effect of parameter T with ASR metric on the target model.
Because we find that T can control the range of semantic appearance deformation as shown in Figure \ref{figure1}.
Figure \ref{figure7} shows the ASR performance when changing T on the FFHQ and CelebA-HQ datasets. 
It can be found that the value of ASR slightly increases with increasing the value of T.
Additionally, Figure \ref{figure1} shows the change of semantic appearance deformation when increasing the value of T.
When the value of T increases, the semantic appearance deformation degree also becomes greater.
We can find more generated adversarial perturbations are integrated into semantic regions (e.g., hair, ear, hat, etc.).
The sufficient experiment results demonstrate that our proposed method guarantees the generated adversarial appearance surrounding the identity-sensitive region and maintains the strong attacking ability.



\subsection{Limitations and Future Work}

Based on the aforementioned analysis, our method has demonstrated the capability to generate high-quality adversarial samples with a high attacking performance. 
However, it is also important to emphasize the limitations of our approach.
Firstly, since our attack model relies on pre-trained large generative models, 
the generation of the adversarial image process is subject to the constraints derived from these source generative models. 
Consequently, some generated adversarial images may exhibit noticeable artifacts, thereby compromising generation quality and potentially impacting the attacking performance.
The negative results are shown in the supplement.
Secondly, it is worth noting that our method primarily focuses on generating adversarial face images, and its ability for other image types may be limited. 
We extend the Adv-Diffusion applied to non-face images in the supplement. 
In the future, we will explore extending the proposed method to encompass a broader range of image types to mimic more real-world scenarios.

\section{Conclusion}

In this paper, we propose the imperceptible adversarial face identity attack algorithm with the latent diffusion model (Adv-Diffusion).
The proposed method designs a novel unified adversarial face image generation framework, which can learn adversarial semantic perturbations in the latent space for high attack capabilities and low perceptibility.
To improve the stealthiness performance, we design the identity-sensitive conditioned diffusion generative module to guarantee the distinct adversarial appearance surrounding the identity-sensitive region.
The adaptive strength-based semantical perturbation is proposed to ensure good stealthiness.
Experiments on the public FFHQ and CelebA-HQ Face datasets illustrate the superior performance of the proposed method.
Noting the proposed method is without any training data or complex network architectures, which will inspire more researchers to explore related fields.
In the future, we will evaluate the proposed method with the identity-attacking performance on more complex real scenarios to adapt to the needs of the real world.

\section{Acknowledgments}
This work was supported in part by the National Natural Science Foundation of China under Grant 62306227, Grant 62276198, Grant U22A2035, Grant U22A2096, Grant 62036007, Grant 61922066, Grant 61906143, Grant 61902297, Grant 61806152 and Grant 61876142; in part by the Key Research and Development Program of Shaanxi (Program No. 2023-YBGY-231); in part by Young Elite Scientists Sponsorship Program by CAST under Grant 2022QNRC001; in part by the Guangxi Natural Science Foundation Program under Grant 2021GXNSFDA075011; in part by Natural Science Basic Research Plan in Shaanxi Province of China under Grant 2022JQ-696; in part by the Technology Innovation Leading Program of Shaanxi under Grant 2022QFY01-15; in part by CCF-Tencent Open Fund, in part by Open Research Projects of Zhejiang Lab under Grant 2021KG0AB01, and in part by the Fundamental Research Funds for the Central Universities under Grant QTZX23083, Grant XJS221502 and Grant QTZX23042.

\bibliography{ref}

\clearpage
\setcounter{figure}{0}
\setcounter{table}{0}
\appendix 


\section{Additional Experimental Details}

In this section, we will introduce more detailed experiment settings of our proposed Adv-Diffusion and other competitors. The mentioned metric is also further clarified clearly here.

\paragraph{Adv-Diffusion.}
We implement the proposed Adv-Diffusion framework on the platform Pytorch.
The latent diffusion model in our method is based on the official code of stable-diffusion. 
Specifically, the official checkpoint for the image inpainting task is chosen as the pre-trained latent diffusion model in the work for better generation performance.
\paragraph{Competitors Details.}
In comparison experiments, we implement FGSM, PGD, MI-FGSM and AMT-GAN to evaluate attack capabilities on public FFHQ and CelebA-HQ datasets.
For FGSM, PGD and MI-FGSM, our code is based on public Torchattacks, which is a PyTorch library that contains adversarial attacks to generate adversarial examples.
During the evaluation, we set the maximum perturbation $\epsilon$ to 8/255 and the maximum iteration to 10 if applicable.
For AMT-GAN, we re-implement it following the official guidance on CelebA-HQ and FFHQ datasets.
For a fair comparison, the experimental setting is the same as the source paper.

\paragraph{ASR Metric.}
In our experiments, we calculate ASR with
FAR@0.01 follows the same protocol in AMT-GAN.
Thus, the parameter $\tau$ of each target model will be set to 0.241, 0.167, 0.409 and 0.302 for IRSE50, IR152, Facenet and Mobileface respectively following Adv-Makeup for a fair comparison.

\paragraph{Computation Burden.}
The computational burden of the method is mainly related to the inference process. 
We conduct experiments on RTX 3090 GPU with a batch size of $2$, and each batch consumes about $62.1s$.

\section{More Analysis of Semantic Appearance Deformation Degree}

\subsection{The effect of attack transferability}

In this section, we further explore the effect of semantic appearance deformation degree.
As illustrated in the paper, the value of $T$ can affect the degree of semantic appearance deformation, which also plays an important role in balance attack transferability and stealthiness.
We separately set the value of $T$ as 50 and 1000 to evaluate the attack ability as shown in Table \ref{ASR_results_1} and Table \ref{ASR_results_2}.

As mentioned in the paper, the value of the semantic appearance deformation degree can balance the attack ability and stealthiness.
In other words, the higher value can bring in better attack ability, while the quality of adversarial faces will decrease.
Experimental results prove the proposed Adv-Diffusion can achieve satisfactory attack performance with suitable appearance deformation.

\begin{table}[htbp]
	\caption{Experimental results with ASR metric on the CelebA-HQ dataset.}
	\centering
	\renewcommand\arraystretch{1.1}        
	\resizebox{\linewidth}{!}{
		\begin{tabular}{*{5}{c}}
			\toprule
			Dataset & \multicolumn{4}{c}{CelebA-HQ}\\
			\cmidrule(lr){1-1} \cmidrule(lr){2-5}
			Target Model & IR152 & IRSE50 & FaceNet & MobileFace \\
			\midrule
			Adv-Diffusion (T=50) & 53.33 & 84.11 & 36.84 & 73.02 \\
			Adv-Diffusion (T=1000) & 56.71 & 88.45 & 46.63& 78.91 \\
			\bottomrule
	\end{tabular}}
	\label{ASR_results_1}
\end{table}

\begin{table}[htbp]
	\caption{Experimental results with ASR metric on the FFHQ dataset.}
	\centering
	\renewcommand\arraystretch{1.1}        
	\resizebox{\linewidth}{!}{
		\begin{tabular}{*{5}{c}}
			\toprule
			Dataset & \multicolumn{4}{c}{FFHQ}\\
			\cmidrule(lr){1-1} \cmidrule(lr){2-5}
			Target Model & IR152 & IRSE50 & FaceNet & MobileFace \\
			\midrule
			Adv-Diffusion (T=50) & 53.42 & 82.91 & 32.21 & 69.25 \\
			Adv-Diffusion (T=1000) & 56.90 & 87.73 & 40.41 & 76.84 \\
			\bottomrule
	\end{tabular}}
	\label{ASR_results_2}
\end{table}

\subsection{The effect of attack stealthiness}

\begin{figure*}[htbp]
	\centering
	\includegraphics[width=0.8\textwidth]{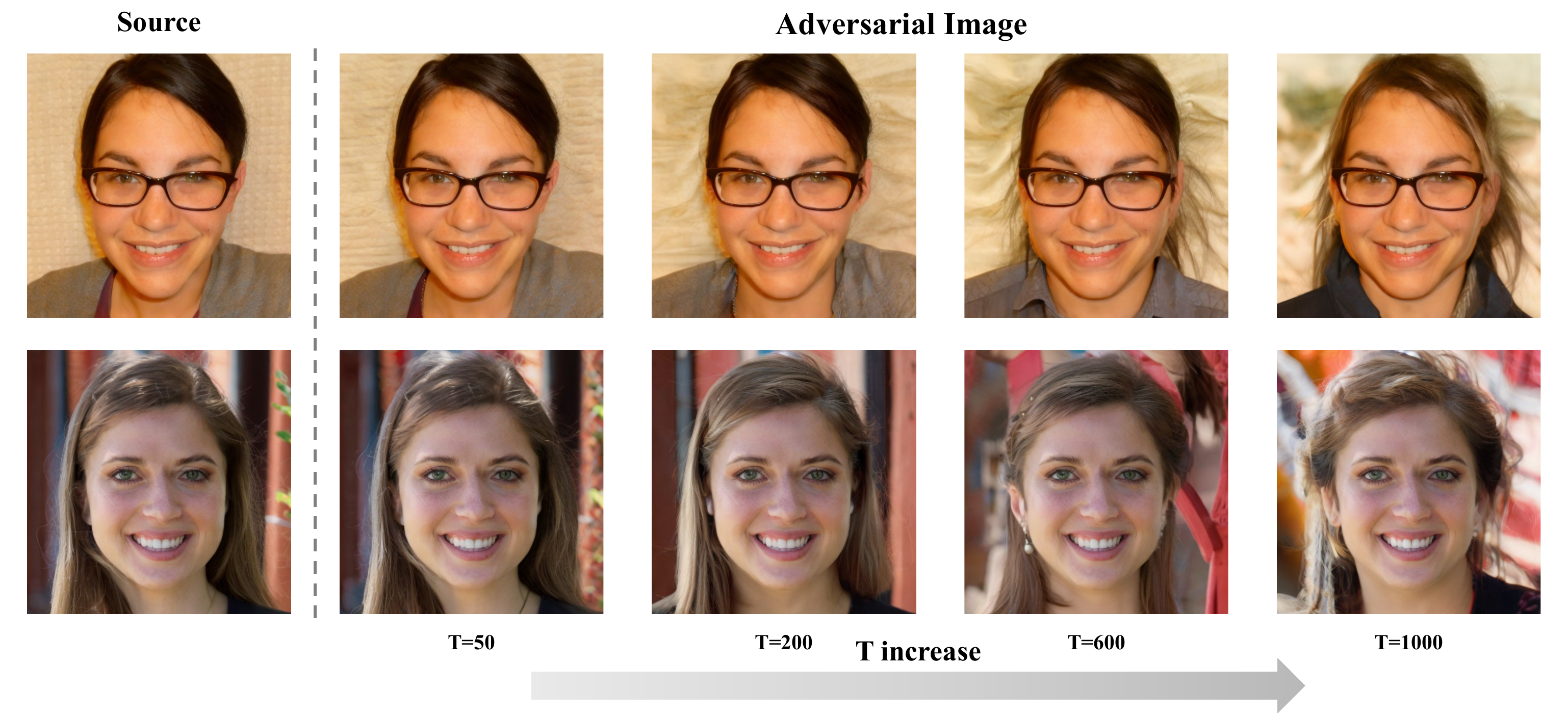}
	\caption{Range of Semantic Appearance Deformation.}
	\label{a_figure1}
\end{figure*}

Furthermore, we explore the semantic appearance changes in the generated images with $T$ value changes. 
As shown in Figure \ref{a_figure1}, When increasing the value of $T$, the semantic appearance deformation changes in the generated images become greater, and more perturbations integrate into semantic regions (e.g. hair, ear, hat, etc,).
The semantic appearance changes greater in identity-agnostic regions.
Similarly, the same phenomenon can be found in Table \ref{a_table3} with image quality metric FID.
It is because FID can effectively quantify the realism and diversity of images generated by generative networks.
Additionally, Figure \ref{a_figure1} demonstrates that the proposed method guarantees the most adversarial appearance surrounding the identity-sensitive region and maintains the central region.
Thus, the adversarial faces generated with ours can conclude suitable semantic adversarial perturbation, and achieve a high ASR performance while ensuring image quality.
In real-world scenarios, we can adjust the value of $T$ according to  specific requirements for the semantic appearance changes of generated adversarial face images.

\begin{table}[htbp]
	\caption{Image quality results on the FFHQ dataset}
	\centering
	\begin{tabular}{*{4}{c}}
		\toprule
		Dataset & \multicolumn{3}{c}{FFHQ} \\
		\cmidrule(lr){1-1} \cmidrule(lr){2-4}
		Metric & FID & PSNR & SSIM \\
		\midrule
		FGSM & 105.68 & 29.52 & 0.82 \\
		PGD  & 73.81 & \textbf{31.89} & \textbf{0.86} \\
		MIFGSM & 77.70 & 30.78 & 0.83\\
		AMT-GAN & 24.54 & 9.33 & 0.37\\
		Adv-Attribute* & 74.86 & -&-\\
		\midrule
		Adv-Diffusion (T=50) & \textbf{18.10} & 27.54& 0.79 \\
		Adv-Diffusion (T=1000) & 45.51 & 14.40 & 0.53 \\
		\bottomrule
	\end{tabular}
	\label{a_table3}
\end{table}
\section{Additional Experimental Analysis}

\subsection{Non-face image application}

Our method can be applied to non-face images for classification tasks conveniently. We implement the Adv-Diffusion method for images from the Internet. Here we use ResNet50
as the target model and directly input source images without
masks to the latent encoder to get the condition c. As shown
in the following figure \ref{figure_non}, the generated adversarial samples
contain both good attacking transferability and stealthiness
when evaluated with ResNet101.

\begin{figure}[htbp]
	\centering
	\includegraphics[width=0.45\textwidth]{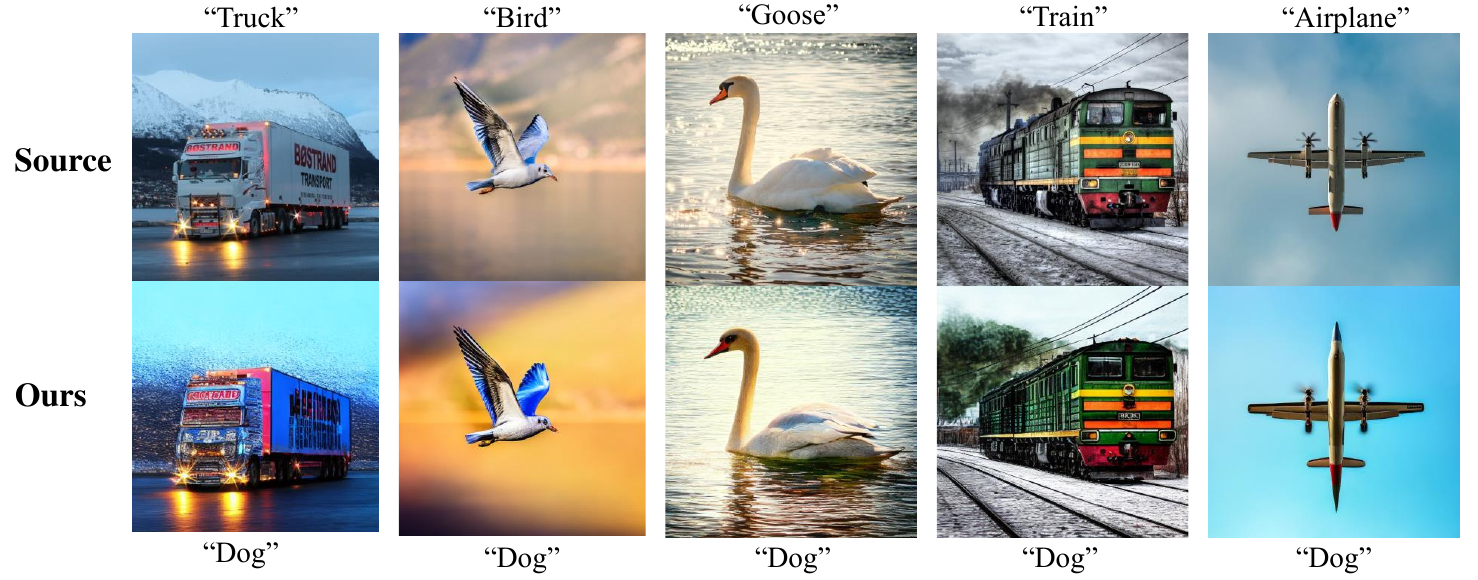}
	\caption{The non-face image results of the proposed adversarial face generation algorithm.}
	\label{figure_non}
	
\end{figure}

\subsection{Face generation in the wild}

We implement the proposed method for face images in the wild with large pose
angles and different backgrounds as shown in the following figure \ref{figure_wild}. Experimental results prove the proposed method
can be suitable in most real-world scenarios. For extremely
complex scenes, we can leverage more advanced parsing
models (or manually tuning) to improve the generation quality. Besides, the extended method can alleviate the effect of
parsing performance, further proving the potential attacking performance in real applications.
\begin{figure}[htbp]
	\centering
	\includegraphics[width=0.45\textwidth]{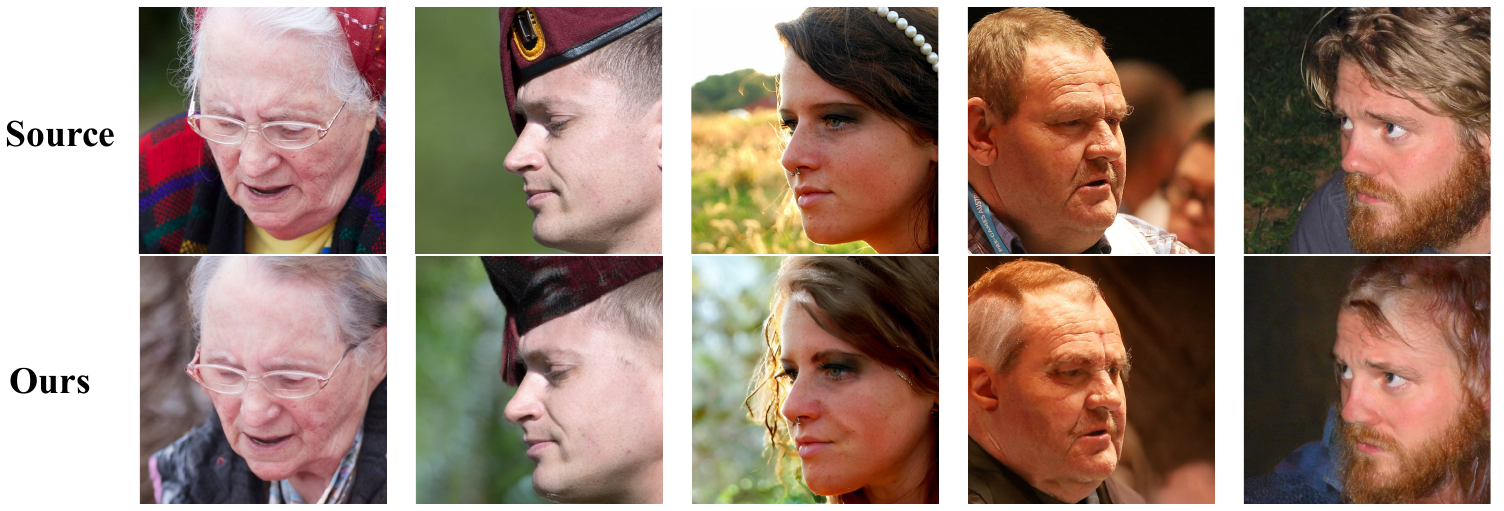}
	\caption{The face in the wild results of the proposed adversarial face generation algorithm.}
	\label{figure_wild}
\end{figure}

\subsection{Failure case}
We add some negative results as shown in the following figure \ref{figure_failure}.
We find that complex hair styling is more likely to bring adverse effects of the parsing and diffusion generative models.
Thus, we will explore a more robust adversarial generative
pipeline in future work.
\begin{figure}[htbp]
	\centering
	\includegraphics[width=0.45\textwidth]{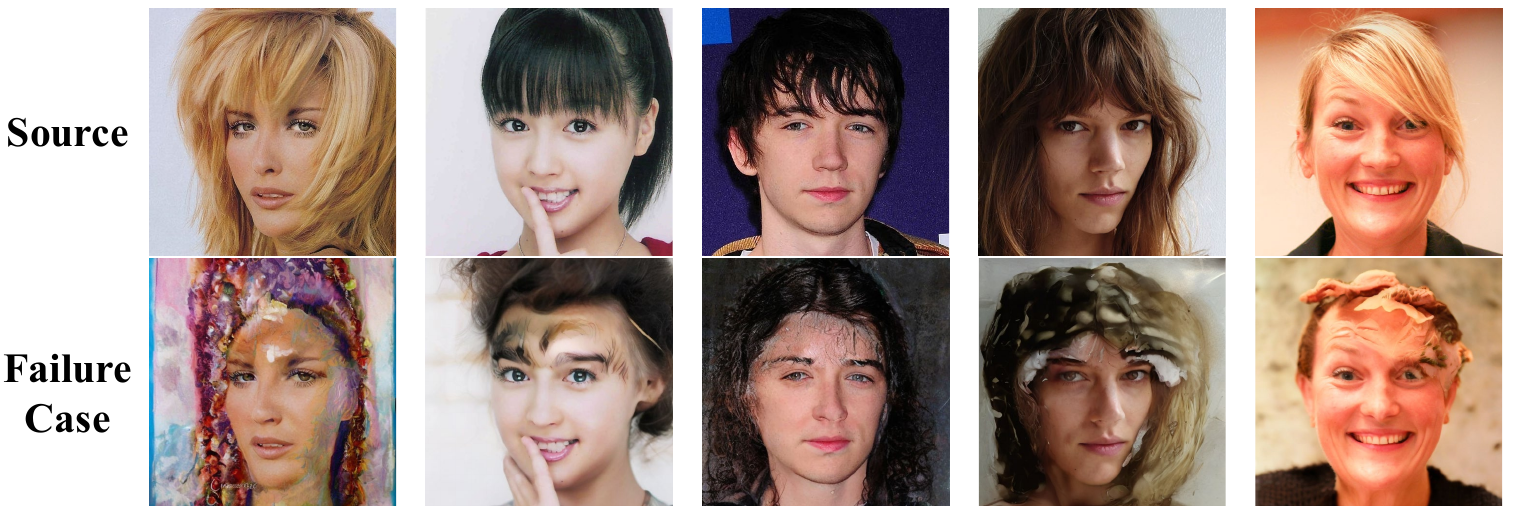}
	\caption{The failure generation results of the proposed adversarial face generation algorithm.}
	\label{figure_failure}
\end{figure}

\subsection{User study}
We conduct a series of user studies with the Visual Turing Test type
to evaluate the performance of Adv-Diffusion, in contrast
to existing MI-FGSM and AMT-GAN methods. In particular, we have 27 participants, all of whom are not professional artists. For each participator, we show them 40 group
randomly selected adversarial face images. Each time, we
show a source image and 3 adversarial generated images
generated by different methods. Participants are requested to
choose the best adversarial images, according to (1) guessing if any attack is injected and (2) the quality of a face
image based on their own preference. Finally, we collect
all preference labels and find 65.6\% participants who think
our model can generate the best adversarial face images,
11.8\% and 22.6\% participants separately voted for AMT-
GAN and MI-FGSM methods. The subjective comparison
results demonstrate ours outperforms existing methods.

\section{Visualization Comparison and Analysis}

\begin{figure*}[htbp]
	\centering
	\includegraphics[width=\textwidth]{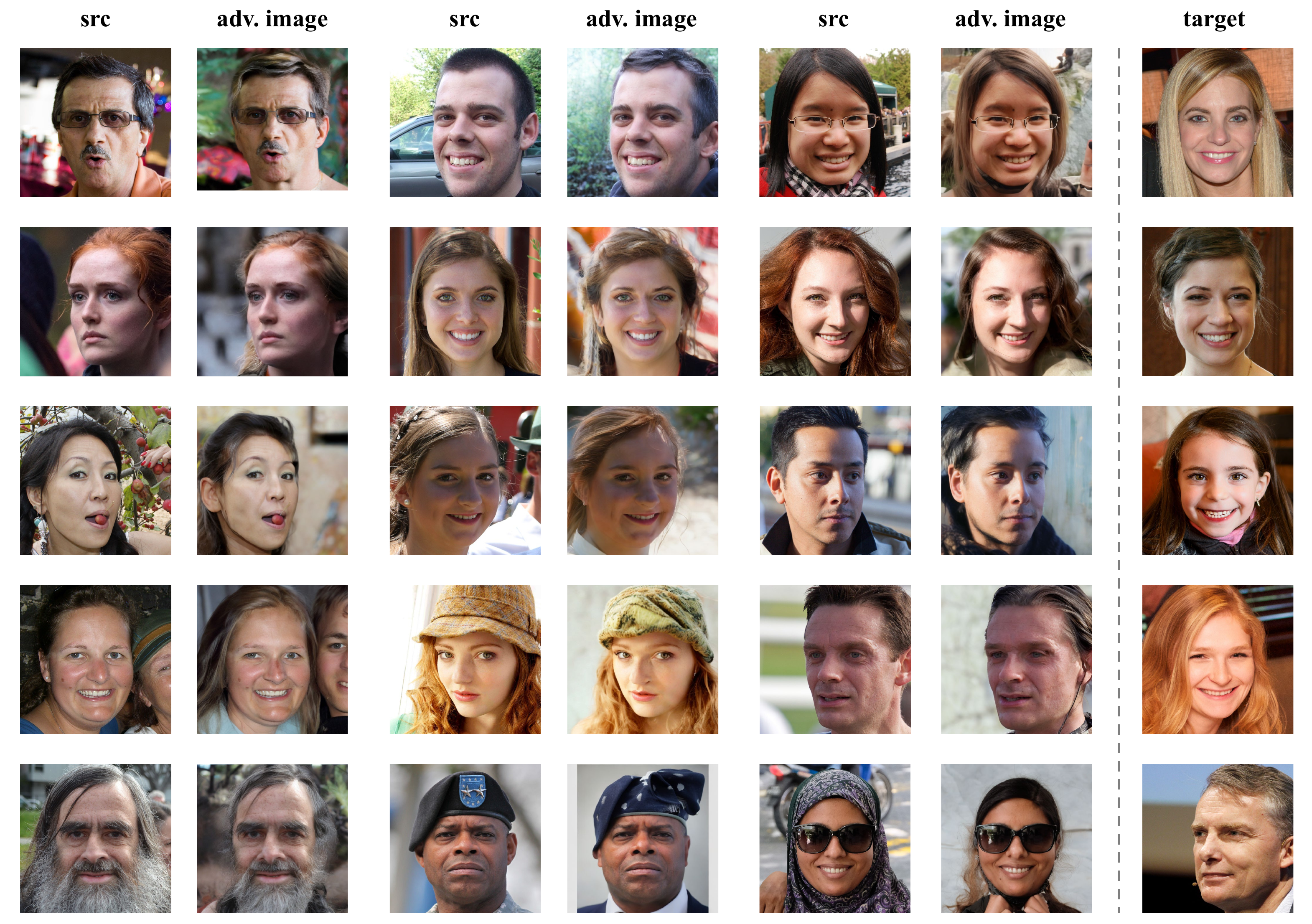}
	\caption{More face synthesis results of the proposed adversarial face generation algorithm.}
	\label{a_figure2}
\end{figure*}

The additional generated images of our proposed Adv-Diffusion are shown in Figure \ref{a_figure2}.
We find that our proposed Adv-Diffusion can generate high-quality adversarial images with imperceptible perturbations.
The generated images' identities are almost the same as the original images, which is hard to be distinguished by human eyes.

It is noted there still exit some failure cases, which are shown in the first column of the last row of Figure \ref{a_figure2}.
The generated adversarial faces seem not realistic and the generated hair is weird.
It is probably because the background is complex and may affect the synthesis quality.
In the future, we will explore a stronger adversarial face generation network with better generalization in real-world scenarios.

\end{document}